\pdfoutput=1
\documentclass[lettersize,journal]{IEEEtran}

\usepackage{amsmath,amsfonts}
\usepackage{array}
\usepackage[caption=false,font=normalsize,labelfont=sf,textfont=sf]{subfig}
\usepackage{graphicx}
\usepackage{stfloats}
\usepackage{balance}
\usepackage{placeins}
\usepackage{float}
\usepackage{capt-of}
\usepackage{textcomp}
\usepackage{url}
\usepackage{verbatim}
\usepackage{booktabs}
\usepackage{multirow}
\usepackage{makecell}
\usepackage{cite}

\usepackage[table]{xcolor}
\definecolor{gainGreen}{RGB}{0,128,64}
\newcommand{\gain}[1]{\textcolor{gainGreen}{(\raisebox{0.2ex}{\scalebox{0.65}{$\uparrow$}}\textbf{#1})}}
\newcommand{\paraminc}[1]{\textcolor{gainGreen}{(+\textbf{#1})}}

\begin{document}

\title{Boundary-Aligned Contribution Routing for Robust Optical--SAR Object Detection}


\author{Haifa~Zhang,
        Yijing~Wang,
        Haoyu~Wang,
        Zheng~Li,
        and~Zhiqiang~Zuo,~\IEEEmembership{Senior~Member,~IEEE}%
        
\thanks{This work was supported in part by the National Natural Science Foundation of China under Grants 62403348 and 62403350; in part by the Postdoctoral Fellowship Program of CPSF under Grant 2026T190468; in part by the Foundation of Key Laboratory of System Control and Information Processing, Ministry of Education, P.R. China, under Grant Scip20240116; and in part by the Emerging Frontiers Cultivation Program of Tianjin University Interdisciplinary Center. (Corresponding author: Haoyu Wang.)}

\thanks{The authors are with the Tianjin Key Laboratory of Intelligent Unmanned Swarm Technology and System, School of Electrical and Information Engineering, Tianjin University, Tianjin 300072, China (e-mail: zhanghaifa@tju.edu.cn; yjwang@tju.edu.cn; why2014@tju.edu.cn; zhengl@tju.edu.cn; zqzuo@tju.edu.cn). H. Wang is also with the Key Laboratory of System Control and Information Processing, Ministry of Education of China, Shanghai 200240, China (e-mail: why2014@tju.edu.cn).}%
}

\markboth{}{}

\maketitle

\begin{abstract}

Optical imagery provides rich appearance cues, whereas synthetic aperture radar (SAR) offers observations that are less sensitive to illumination and weather, making optical--SAR fusion attractive for remote-sensing object detection. However, the presence of multiple modalities does not guarantee beneficial fusion: imperfect spatial, temporal, and semantic correspondence can make an otherwise intact stream conditionally harmful and induce negative cross-modal transfer. We handle this issue through a model-specific task-utility perspective and learn task-conditioned contribution routing using detection supervision alone. The proposed fusion-boundary-aligned routing regulates each modality's contribution before the first learned cross-modal feature-value mixing operation. For architectures with frequent shallow interaction, a Feature Router performs cross-conditioned, group-addressable modulation near the input; for dual-backbone architectures, a Dual-Statistic Semantic Router predicts stream-level contribution weights from modality-specific average and maximum statistics before late semantic fusion. The routers require no explicit utility supervision, quality labels, reconstruction, or distillation. Experiments on M4-SAR and SpaceNet6-OTD cover nominal full inputs, controlled correspondence shifts, missing modalities, and four nonzero modality-corruption scenarios. Across the reported clean-training controls, routing improves full-input $\text{mAP}_{50}$ by 0.5--5.9 points. Relative to the corresponding modality-dropout baselines, it raises missing-modality $\text{mAP}_{50}$ by 7.6--41.6 points and reduces the negative-transfer rate by up to 12.7 percentage points. Spearman correlations between the learned routing weights and model-specific leave-one-modality-out utility range from 0.45 to 0.66, supporting the task-utility interpretation of the routing coefficients.
\end{abstract}


\section{Introduction}
\label{sec:intro}

\IEEEPARstart{M}{ultimodal} object detection exploits the complementary sensing characteristics of Optical and Synthetic Aperture Radar (SAR) imagery~\cite{25DEN, lin2021low, ye2022unsupervised}. Optical imagery provides interpretable texture and appearance cues, whereas active microwave sensing remains informative under adverse illumination and weather but exhibits speckle and weaker visual semantics. These complementary properties motivate Optical--SAR fusion for Earth observation~\cite{rajah2018feature, xu2026nam}.

\begin{figure}[t]
\centering
\includegraphics[width=0.90\columnwidth]{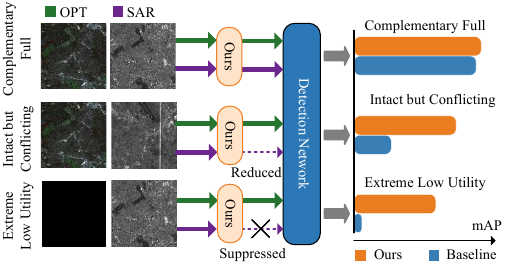}
\caption{Motivation for fusion-boundary-aligned contribution routing. A present and visually valid stream can still conflict with the other modality and cause negative transfer. We control source-indexed contributions before the first learned cross-modal feature-value mixing operation.}
\label{fig:intro_motivation}
\end{figure}

Complementarity, however, is not equivalent to universal usefulness. Optical--SAR pairs rarely achieve satisfactory spatial, temporal, scale, and semantic correspondence due to registration error, resolution mismatch, acquisition-time gaps, heterogeneous sensing mechanisms, and scene change. Consequently, a visually intact modality can still introduce conflicting features and degrade detection accuracy after fusion. We therefore regard modality usefulness as a task-conditioned and interaction-dependent quantity rather than an intrinsic image-quality score. As illustrated in Fig.~\ref{fig:intro_motivation}, this variable-utility spectrum ranges from complementary full inputs, through intact-but-conflicting pairs, to extreme low-utility cases such as missing or visibly corrupted modalities, motivating routing that retains, reduces, or suppresses source contributions before the first learned cross-modal feature-value mixing operation.

Modern fusion architectures include early concatenation~\cite{ngiam2011multimodal,10891164}, cross-modal attention~\cite{qingyun2022cross,SHEN2023109913}, and state-space models~\cite{25M3,25MaDiNet,26MMMF}. It is noted that most of them emphasize stronger interaction but do not ask whether every source feature should enter the joint transformation for the current sample. When low-utility or conflicting content is mixed without contribution control, it will cause negative cross-modal transfer. This observation raises two questions: how should modality contribution be represented, and at what point should it be controlled before harmful content is mixed into a shared representation?

Existing robust multimodal methods primarily treat absence or visible corruption through reconstruction and knowledge distillation. Generative approaches recover missing information from cross-modal correlations~\cite{yeh2017semantic,guan2022deepmih,zhou2021feature}, whereas distillation transfers full-modality knowledge to incomplete-input models~\cite{gupta2016cross,shu2021channel,zhuang2025cmad,chen2024novel}. These methods address important boundary cases but do not directly model sample-dependent contribution under nominal full inputs. Their auxiliary reconstruction or alignment objectives can also diverge from detection utility.

Here we propose \emph{fusion-boundary-aligned contribution routing} from this model-specific task-utility perspective. The implementation follows fusion topology rather than imposing one module everywhere. Architectures with frequent shallow interaction use an input Feature Router (FR): modality-specific scalar weights precede concatenation, after which cross-conditioned channel and spatial gates modulate the still-addressable Optical/SAR groups. Dual-backbone architectures use a Dual-Statistic Semantic Router (SR), which maps average and maximum statistics of each separated high-level stream to a contribution weight before late fusion. FR and SR are mutually exclusive realizations of the same placement principle. The detection objective learns the weights without quality labels, reconstruction, distillation, or explicit utility supervision. And the leave-one-modality-out (LOMO) utility remains an analytical diagnostic rather than the router's optimization target; modality dropout only broadens the observed range of input states.

The proposed routing is therefore not restricted to unavailable or visibly corrupted inputs. Under nominal full-modality observations, imperfect correspondence can make one stream conditionally harmful. Boundary-aligned routing mitigates this negative transfer and can improve full-input detection beyond an ordinary clean-trained fusion baseline.

The main contributions of this work are summarized as follows:
\begin{itemize}
    \item We propose a model-specific task-utility formulation for multimodal fusion via detector-level counterfactual analysis instead of modality-quality assessment, and employ LOMO utility as an analytical diagnostic of sample-dependent contribution rather than an optimization target.

    \item We identify the first learned cross-modal feature-value mixing boundary and propose topology-aligned FR/SR routing that controls contribution while source-indexed feature-value paths remain explicitly addressable.

    \item We achieve state-of-the-art performance under missing modalities among the compared methods on M4-SAR and SpaceNet6-OTD, with a maximum $\mathrm{mAP}_{50}$ improvement of 41.6 points. Controlled shifts, LOMO-utility correlations, and negative-transfer analysis further validate full-input enhancement, boundary placement, and the empirical utility interpretation.
\end{itemize}

\section{Related Work}
\label{sec:related_work}
\subsection{Multimodal Remote Sensing Perception}
\label{subsec:multimodal_rs}

The fusion of optical and SAR imagery has been extensively investigated for robust remote sensing perception, owing to the complementary sensing characteristics of these modalities~\cite{li2022multimodalrsreview}. Optical imagery provides rich appearance and texture cues, whereas SAR is more resilient to illumination and weather variations but typically contains weaker semantic details and is affected by speckle noise. Such complementarity has supported optical-enhanced object detection and heterogeneous time-series change monitoring~\cite{Zhang22otd,Wang2023FloodChange,zhang2023superyolo}.

At the image level, complementary-feature decomposition and visual saliency have been utilized to combine optical and SAR observations~\cite{Ye2024Complementary,25DEN}. Related hyperspectral--multispectral fusion methods employ deep spectral--spatial constraints or self-supervised unrolling to reconstruct high-resolution products~\cite{Yu2024SpectralSpatial,Zhu2024Unrolling,li2022multimodalrsreview}.

Existing Optical--SAR fusion schemes for perception can generally be categorized into early-fusion and late-fusion paradigms~\cite{baltrusaitis2019multimodal,li2022multimodalrsreview}. Conventional approaches primarily rely on early feature concatenation~\cite{10891164} or late decision-level fusion~\cite{eitel2015multimodal,10438483} to integrate cross-modal information. More recent deep learning-based developments have introduced cross-modal attention, dynamic interaction, frequency-domain decoupling, state space modeling, and bidirectional distillation to enhance heterogeneous feature interaction and alignment~\cite{xu2023multimodaltransformers,bao2025dualdynamic,yue2025diffusionkd,25M3,26MMMF,25MaDiNet,Sun2024FrequencyExperts}. Dedicated affine and deformable registration networks can reduce geometric discrepancies before fusion~\cite{Xiao2024ADRNet}, but do not directly control sample-wise task contribution.

Most fusion results depend on the premise that additional modality content is beneficial whenever it is available~\cite{baltrusaitis2019multimodal,rahate2022multimodal}. This assumption may fail in the absence of visible degradation: imperfect correspondence may render intact source features low-utility or conflicting for the joint detector. Missing inputs, clouds, and speckle constitute severe manifestations of the broader variable-utility problem~\cite{ma2022multimodal,xu2026nam,chen2024novel}.

\subsection{Cross-Modal Conflict and Reliability-Aware Fusion}
Prior work has recognized that multimodal interaction can be unreliable. Conflict-aware multispectral detection explicitly addresses disagreement between visible and infrared features~\cite{he2023multispectral}, while channel-selection and spatial-attention fusion adaptively govern which features are combined~\cite{23CSSA,24MMIDet}. A related line estimates predictive confidence or view uncertainty: confidence-aware fusion and CML regulate uncertain predictions~\cite{24MCHE,ma2023cml}, while trusted multi-view and dual-level evidential fusion perform uncertainty-aware integration~\cite{23Trusted,shao2024dualevidential}. These findings indicate that, rather than fusing modalities with uniform weighting, fusion control variables should capture feature attention, predictive confidence, or evidential uncertainty.

Although these treatments improve conflict awareness, their learned variables are usually tied to attention, confidence, or uncertainty~\cite{24MCHE,23Trusted,shao2024dualevidential}. And they do not directly quantify the counterfactual contribution of each modality to a fixed detector, nor do they examine whether the location of a gate relative to the first cross-modal feature-value mixing operation affects its ability to prevent negative transfer.

\subsection{Missing Modalities and Robust Perception}
\label{subsec:missing_modalities}

Robust multimodal perception with missing or corrupted modalities has received increasing attention because sensor failure and severe local degradation reduce downstream accuracy~\cite{rahate2022multimodal,ma2022multimodal,wang2023multi,zhang2024tmformer,Park2025Resilient,BLUM2005119,chen2024novel,23MSH}. Existing approaches are broadly divided into reconstruction-based and distillation-based streams~\cite{rahate2022multimodal}.

Reconstruction-based methods recover missing views or features from cross-modal dependencies~\cite{Zhang2022MMFormer,yang2024incomplete,Dong2024MissingOptical,chen2024novel}, while distillation-based ones transfer full-modality knowledge to a student operating under incomplete inputs~\cite{liu2023m3ae,liu25Reliable,liu2024modality,23MSH}. These compensation strategies will incur overhead and rely on proxy objectives, and their primary focus remains complete modality absence rather than sample-dependent contribution under nominal full inputs~\cite{rahate2022multimodal}.

Recent robustness-oriented detectors increasingly employ architecture-specific alignment, Transformer, quality-token, dynamic-interaction, knowledge-distillation, or expert-routing components~\cite{Shi2026BANet,Zhao2026QDFNet,yan2023cross,Park2025Resilient,xu2023multimodaltransformers,bao2025dualdynamic,yue2025diffusionkd}. Although effective, these designs may introduce substantial parameter or computation overhead, particularly in Transformer- and multi-expert variants, and often require nontrivial redesign for different fusion topologies. Moreover, their learned quality or routing variables are not consistently related to detector-specific counterfactual utility, which limits their task-level interpretability.

\subsection{Feature Gating and Routing Mechanisms}
\label{subsec:feature_gating}

Feature gating and routing mechanisms, including Squeeze-and-Excitation (SE) networks~\cite{hu2018squeeze} and Mixture-of-Experts architectures~\cite{riquelme2021scaling}, dynamically modulate feature propagation~\cite{han2022dynamic}. Multimodal variants often interpret gates through availability, confidence, or input quality~\cite{rahate2022multimodal,shao2024dualevidential}. Their placement is commonly selected as an architectural choice rather than analyzed relative to the point at which modality-indexed feature values first become irreversibly mixed.

However, several existing gating modules rely primarily on global average pooling~\cite{hu2018squeeze,han2022dynamic}. A single statistic would be insufficient when strong local responses arise from either targets or degradation artifacts. Complementary average and maximum descriptors provide a richer input for task-conditioned routing, although they do not by themselves constitute calibrated reliability or uncertainty estimates~\cite{23Trusted,shao2024dualevidential}.

\subsection{Summary}
Existing studies provide strong mechanisms for heterogeneous feature interaction, conflict awareness, missing-modality compensation, and dynamic gating~\cite{li2022multimodalrsreview,rahate2022multimodal,han2022dynamic}. Nevertheless, three issues are not fully explored: sample-specific contribution to the detector is rarely quantified counterfactually; nominal full inputs and missing or corrupted inputs are usually treated as separate robustness regimes; and routing position is seldom defined relative to the first learned cross-modal feature-value mixing operation. We address these issues by using model-specific LOMO utility only as an analytical diagnostic, defining a feature-value mixing boundary, and deploying topology-aligned FR or SR modules before that boundary. The routers are optimized by detection supervision without utility, quality, reconstruction, or distillation labels.

\section{Methodology}
\label{sec:methodology}

\subsection{Task Utility and Feature-Value Mixing Boundary}
Let $(x_o,x_s,y)$ be an Optical image, a SAR image, and the detection target, respectively. For subsequent analysis, we quantify the counterfactual contribution of modality $m\in\{o,s\}$ via model-specific leave-one-modality-out (LOMO) task utility, where $o$ and $s$ index the Optical and SAR streams. Before cross-modal interaction, these two streams are represented as
\begin{equation}
h_o^l=E_o^l(x_o),\qquad h_s^l=E_s^l(x_s),
\label{eq:separate_streams}
\end{equation}
where $E_m^l$ is the modality-specific encoder mapping at layer $l$, and $h_m^l$ is its output feature representation.

The LOMO task utility is the loss reduction obtained by adding modality $m$ to the complementary stream:
\begin{equation}
u_m=\mathcal L_{det}\!\left(D(h_{-m}),y\right)
-\mathcal L_{det}\!\left(D(h_o,h_s),y\right),
\label{eq:oracle_utility}
\end{equation}
where $D$ is the frozen detector, $\mathcal L_{det}$ is its detection loss, and $h_{-m}$ replaces stream $m$ with the zero-valued missing-stream representation in our availability tests. Both terms adopt the same checkpoint and target $y$, without retraining or changing any model parameter. Thus, $u_m$ is a counterfactual property of a particular detector under this intervention, not an intrinsic property of modality $m$. A positive $u_m$ stands for useful contribution, $u_m\approx0$ represents redundancy, and $u_m<0$ means that adding modality $m$ increases the detection loss and causes negative transfer. Across a dataset, we define the \emph{negative-transfer rate (NTR)} as the frequency at which full fusion underperforms either single-stream counterfactual from the same frozen detector:
\begin{equation}
\mathrm{NTR}=\frac{1}{N}\sum_{i=1}^{N}\mathbf 1\!\left[
\mathcal L_i^{full}>\min\!\left(\mathcal L_i^{opt},\mathcal L_i^{sar}\right)
\right],
\label{eq:ntr}
\end{equation}
where $N$ is the number of evaluated samples, $\mathbf 1[\cdot]$ is the indicator function, and $\mathcal L_i^{full}$, $\mathcal L_i^{opt}$, and $\mathcal L_i^{sar}$ are the detection losses of the full, Optical-only, and SAR-only inputs for sample $i$, respectively.
LOMO utility is an analytical diagnostic rather than the optimization target of the router. These counterfactual quantities are used exclusively for analysis. The router receives no utility label, and detection supervision alone justifies interpreting its output solely as a task-conditioned contribution weight. Whether that weight tracks LOMO utility is an empirical question evaluated later by rank correlation; it is not guaranteed by the optimization objective. The routing coefficient $a_m$ does not estimate intrinsic image quality, sensor reliability, or uncertainty. A visually intact stream may nevertheless be downweighted when imperfect registration, temporal inconsistency, or heterogeneous sensing renders its contribution detrimental to the complete detector.

Before defining the boundary, we distinguish control-path interaction from cross-modal feature-value mixing. A learned gate may be cross-conditioned on both streams,
\begin{equation}
\tilde h_m=a_m(h_o,h_s)\odot h_m,\qquad m\in\{o,s\}.
\label{eq:addressability_modulation}
\end{equation}
Here, $a_m(h_o,h_s)$ is the routing coefficient for stream $m$, $\odot$ denotes the Hadamard product, and $\tilde h_m$ is the routed feature. We call Eq.~\eqref{eq:addressability_modulation} \emph{addressability-preserving modulation}: although $a_m$ may read both modalities, every output remains a multiplicative transformation of its original modality-indexed feature. Cross-modal interaction is therefore allowed in the control path. However, no additive or value-projection term injects SAR feature values into the Optical-indexed value path, or vice versa.

Our boundary refers specifically to the first operation that mixes feature values across source-indexed streams, rather than the first operation whose output depends on both modalities. We define \emph{cross-modal feature-value mixing} as a learned transformation that integrates feature values across modality groups, whether by additive combination or other fusion mechanisms (e.g., $W_{o\leftarrow s}h_s$, $W_{s\leftarrow o}h_o$, a joint convolution, or cross-modal attention). The \emph{first learned cross-modal feature-value mixing boundary} $b$ is the earliest such mapping, after which the output no longer admits a fixed channel partition attributable to one input modality:
\begin{equation}
z^b=\Phi_b(h_o^{b-1},h_s^{b-1}),
\label{eq:mixing_boundary}
\end{equation}
where $\Phi_b$ is the first learned cross-modal feature-value mixing operator and $z^b$ represents its output.
Note that simple concatenation is not a feature-value mixing boundary because fixed channel indexing may recover the two groups. A boundary-aligned router instead acts before formula~\eqref{eq:mixing_boundary}:
\begin{equation}
\begin{aligned}
\tilde h_m^{b-1}
&=R_m(h_o^{b-1},h_s^{b-1};\theta_r)\odot h_m^{b-1},\\
z^b
&=\Phi_b(\tilde h_o^{b-1},\tilde h_s^{b-1}).
\end{aligned}
\label{eq:boundary_routing}
\end{equation}
Here, $R_m(\cdot;\theta_r)$ is the router for modality $m$, and $\theta_r$ denotes its trainable parameters.
SR follows this form directly. FR first applies stream-wise scalar routing, concatenates the weighted features, and then performs addressability-preserving channel/spatial modulation. Its gates may read the full concatenated tensor, but their outputs only scale the original indexed channels; the downstream joint transformation is the first feature-value mixing layer.

Training samples a modality-availability state $q$ from
\begin{equation}
\begin{gathered}
\mathcal Q=\{q_{full},q_{-o},q_{-s}\},\\
P(q_{full})=1-2p,\\
P(q_{-o})=P(q_{-s})=p,
\end{gathered}
\label{eq:training_state_distribution}
\end{equation}
where $\mathcal Q$ is the set of allowed availability states, $q_{full}$ retains both streams, $q_{-o}$ masks Optical, $q_{-s}$ masks SAR, and $p=0.2$ is the single-stream masking probability. Thus, full input, Missing Optical, and Missing SAR occur with probabilities $0.6$, $0.2$, and $0.2$, respectively; masking both modalities together is excluded. The training objective is
\begin{equation}
\min_{\theta}\;\mathbb E_{(x_o,x_s,y),\,q\sim\mathcal Q}
\left[\mathcal L_{det}\big(D_{\theta}(x_o^q,x_s^q),y\big)\right],
\label{eq:task_objective}
\end{equation}
where $\theta$ denotes all trainable detector and router parameters, $x_m^q$ is modality $m$ after applying availability state $q$, and $\mathbb E$ denotes expectation over training samples and availability states.

\subsection{Overall Architecture}
\label{subsec:overall_architecture}
\begin{figure*}[t]
\centering
\includegraphics[width=0.90\textwidth]{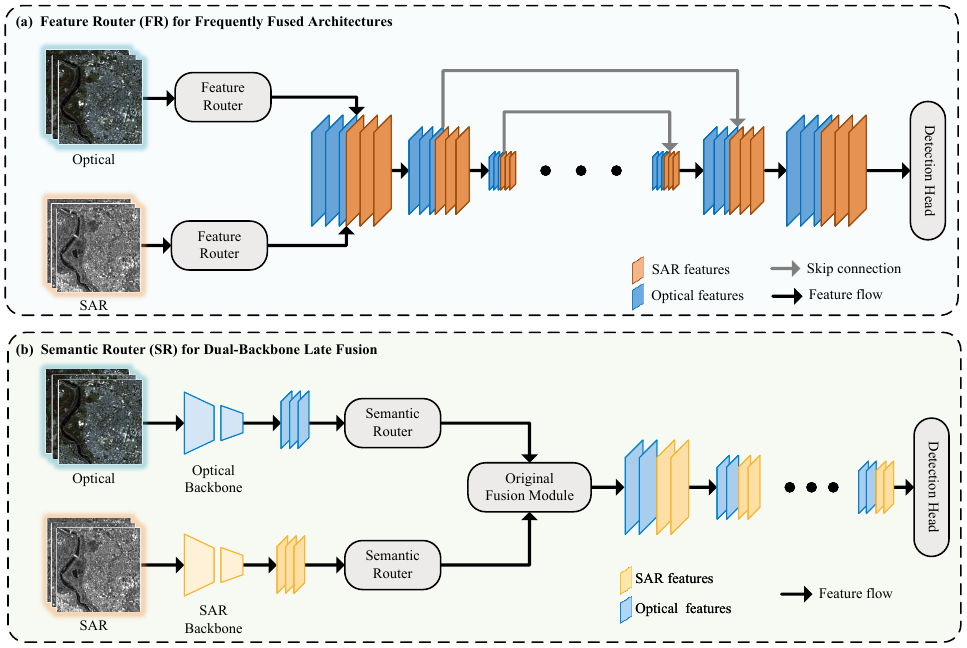}
\caption{Topology-aligned realizations of fusion-boundary contribution routing. FR modulates explicitly addressable input channel groups before the backbone mixes their feature values; SR learns a task-conditioned stream contribution weight from separated semantic features before late fusion. The modules are mutually exclusive choices.}
\label{fig:overall_arch}
\end{figure*}

As illustrated in Fig.~\ref{fig:overall_arch}, the first learned feature-value mixing boundary depends on network topology. In encoder--decoder architectures, the downstream stem learns from the group-addressable FR output, while in dual-backbone frameworks, modality separation persists until semantic fusion and SR is placed immediately before that operation. Actually, FR and SR are alternative topology-matched realizations rather than sequential components.

Specifically, given an input multimodal tensor comprising optical and SAR data, the pipeline operates as follows:

\noindent\textbf{Step 1) Input-boundary routing:} FR computes $w_{opt}$ and $w_{sar}$ before concatenation, then implements group-addressable channel selection ($v_c$) and spatial filtering ($G_{voxel}$). The gates may be cross-conditioned, but their outputs remain multiplicatively tied to the indexed source channels before downstream feature-value mixing.

\noindent\textbf{Step 2) Late-boundary routing:} SR aggregates average-pooled ($F_{ap}$) and max-pooled ($F_{mp}$) responses from each separated semantic stream and generates a task-conditioned contribution weight $\alpha$ before learned cross-modal fusion.

\noindent\textbf{Step 3) Task-driven learning:} The training-state distribution in~\eqref{eq:training_state_distribution} exposes the detector to a wider utility range. The downstream detection loss learns routing weights without quality, reliability, or LOMO-utility labels.

\subsection{Feature Routing Stem}
\label{subsec:fr_stem}

\begin{figure*}[t]
\centering
\includegraphics[width=0.90\textwidth]{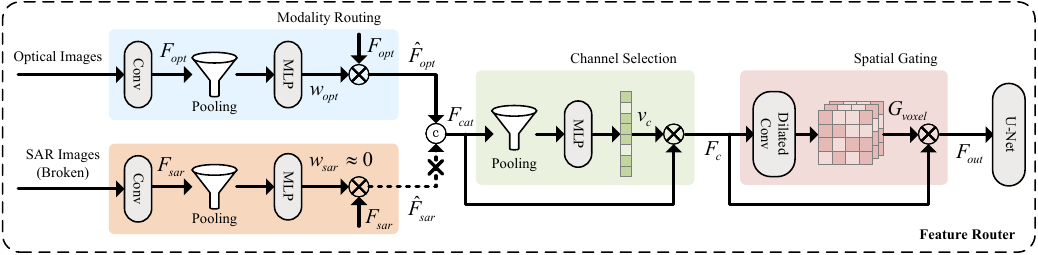}
\caption{Detailed architecture of the FR stem. Modality-specific scalar weighting precedes concatenation; cross-conditioned channel and spatial gates then perform addressability-preserving modulation before downstream feature-value mixing.}
\label{fig:fr_stem}
\end{figure*}

To control low-utility modality effects near the input, we introduce the FR stem as in Fig.~\ref{fig:fr_stem}. FR is not entirely pre-concatenation: its scalar weights are modality-specific, whereas its channel and spatial gates operate on $\mathbf F_{cat}$. Because these gates rescale indexed feature values rather than inject values across source paths, we regard FR as boundary-aligned, group-addressable routing rather than purely pre-fusion routing.

Let $\mathbf{X} \in \mathbb{R}^{B \times 6 \times H \times W}$ be the input multimodal tensor, where $B$ is the batch size, and $H$ and $W$ are the spatial dimensions. The FR stem first splits $\mathbf{X}$ along the channel dimension into two modality-specific components: $\mathbf{X}_{opt} \in \mathbb{R}^{B \times 3 \times H \times W}$ and $\mathbf{X}_{sar} \in \mathbb{R}^{B \times 3 \times H \times W}$.

\subsubsection{Independent Scalar Weighting}
Each modality is processed by an independent scalar branch for preliminary contribution modulation. For each modality $m \in \{opt, sar\}$, the initial feature extraction is formulated as:
\begin{equation}
    F_m = \text{LReLU}(\text{IN}(\text{Conv}_{3\times3}(\mathbf{X}_m))),
\end{equation}
where $\text{Conv}_{3\times3}$ is a $3\times3$ convolution, and $\text{IN}$ and $\text{LReLU}$ are Instance Normalization and Leaky ReLU, respectively. A task-conditioned scalar weight $w_m$ is generated via an SE operation~\cite{hu2018squeeze}:
\begin{equation}
    w_m = \sigma(\mathbf{W}_{m,2} \delta (\mathbf{W}_{m,1} \text{GAP}(F_m))),
\end{equation}
where $\text{GAP}(\cdot)$ is global average pooling, $\delta$ is the ReLU activation, and $\sigma$ is the sigmoid function; $\mathbf{W}_{m,1}$ and $\mathbf{W}_{m,2}$ are learnable weights of the fully connected layers. The weighted modality feature is then $\hat{F}_m = w_m \cdot F_m$.

\subsubsection{Hierarchical Gating Refinement}
The weighted features are concatenated to form a fused representation $\mathbf{F}_{cat} = [\hat{F}_{opt}, \hat{F}_{sar}] \in \mathbb{R}^{B \times C \times H/2 \times W/2}$, where $C$ is the concatenated channel count. To further refine channel-wise and spatial dependencies, a two-stage gating mechanism is proposed:

1) \textit{Channel Importance Selection:} A channel gate $v_c$ recalibrates channel responses:
\begin{equation}
    \mathbf{F}_c = \mathbf{F}_{cat} \otimes v_c.
\end{equation}
The channel gate is computed as $v_c = \sigma(\text{MLP}(\text{GAP}(\mathbf{F}_{cat})))$, where MLP is a multi-layer perceptron and $\otimes$ denotes broadcast element-wise multiplication.

2) \textit{Spatial Gating:} To attenuate localized low-utility responses (e.g., local cloud occlusion), a spatial gate $G_{voxel}$ in terms of dilated convolutions is introduced:
\begin{equation}
    \mathbf{F}_{out} = \mathbf{F}_c \otimes G_{voxel},
\end{equation}
where $G_{voxel} = \sigma(\text{Conv}_{1\times1}(\text{D-Conv}_{3\times3}(\text{Conv}_{1\times1}(\mathbf{F}_c))))$ is the spatial gate and $\text{D-Conv}_{3\times3}$ is a dilated $3\times3$ convolution. The cumulative FR variants are evaluated in the component ablation.
\subsection{Dual-Statistic Semantic Routing}
\label{subsec:semantic_router}

\begin{figure}[t]
\centering
\includegraphics[width=0.90\columnwidth]{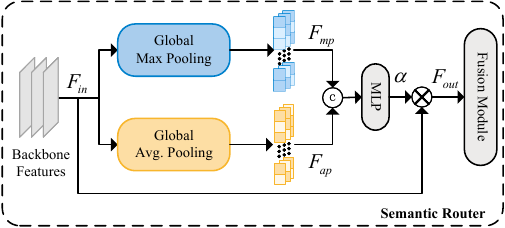}
\caption{Architecture of SR. Average and maximum descriptors are mapped to a task-conditioned routing coefficient; the coefficient is not interpreted as calibrated confidence or uncertainty.}
\label{fig:sr_module}
\end{figure}

For a dual-backbone detector, modality-separated features remain available through the deep stages. We insert SR after the backbones but before their first learned semantic feature-value mixing operation, as shown in Fig.~\ref{fig:sr_module}. SR modulates task contribution; it does not explicitly estimate sensor reliability or geometric alignment.

\subsubsection{Statistical Feature Aggregation}
The SR module operates on the input backbone feature map $F_{in} \in \mathbb{R}^{B \times C \times H \times W}$. Unlike conventional attention mechanisms that rely exclusively on average pooling, the SR aggregates complementary global statistics by concatenating average-pooled and max-pooled responses:
\begin{equation}
    F_{ap} = \text{AvgPool}(F_{in}), \quad F_{mp} = \text{MaxPool}(F_{in}),
\end{equation}
where $F_{ap}, F_{mp} \in \mathbb{R}^{B \times C}$ are the global average and maximum responses across the spatial dimensions, respectively. The average-pooled response reflects the overall activation level, whereas the max-pooled response captures the strongest local activation.

\subsubsection{Task-Conditioned Contribution Gating}
The concatenated global statistics $[F_{ap}, F_{mp}]$ are fed into a lightweight MLP to generate a task-conditioned routing coefficient $\alpha$:
\begin{equation}
    \alpha = \sigma(\mathbf{W}_2 \cdot \delta(\mathbf{W}_1 \cdot [F_{ap}, F_{mp}] + b_1) + b_2),
    \label{eq:sr_full_gate}
\end{equation}
where $\mathbf{W}_1$ and $\mathbf{W}_2$ are learnable weight matrices, and $b_1$ and $b_2$ are their corresponding biases. A reduction ratio is selected for computational efficiency. To avoid suppressing feature transmission during early training, the output bias $b_2$ is initialized to $3.0$, yielding an initial gating value of approximately $0.95$. The final gated feature is $F_{out} = F_{in} \cdot \alpha$, where $F_{out}$ denotes the routed semantic feature.

This coefficient multiplicatively governs feature transmission according to detection supervision. Without calibration or uncertainty labels, $\alpha$ is interpreted only as a routing weight rather than a probability of reliability~\cite{23Trusted}.

\subsubsection{Explicit Ablation Variants}
Pooling produces a descriptor rather than a routing weight. Therefore, every SR ablation maps its descriptor to a scalar gate before feature modulation. Complementing the FR variants, let $\mathbf{w}_{ap}$, $\mathbf{w}_{mp}$, and $\mathbf{w}_{am}$ be learnable linear projections and $b_{ap}$, $b_{mp}$, and $b_{am}$ be their biases. The four SR variants are defined as
\begin{subequations}
\label{eq:sr_ablation_gates}
\begin{align}
\alpha_{ap}^{\mathrm{lin}} &= \sigma(\mathbf{w}_{ap}^{\top}F_{ap}+b_{ap}), \\
\alpha_{mp}^{\mathrm{lin}} &= \sigma(\mathbf{w}_{mp}^{\top}F_{mp}+b_{mp}), \\
\alpha_{am}^{\mathrm{lin}} &= \sigma(\mathbf{w}_{am}^{\top}[F_{ap},F_{mp}]+b_{am}), \\
\alpha_{am}^{\mathrm{mlp}} &= \sigma(\mathbf{W}_2\delta(\mathbf{W}_1[F_{ap},F_{mp}]+b_1)+b_2),
\end{align}
\end{subequations}
where the last expression is the full SR gate in~\eqref{eq:sr_full_gate}. For any variant $q\in\{ap\text{-lin},mp\text{-lin},am\text{-lin},am\text{-mlp}\}$, the routed output is
\begin{equation}
F_{out}^{q}=\alpha^{q}F_{in}.
\label{eq:sr_ablation_output}
\end{equation}
Thus, the first two variants test the descriptor statistic while keeping a linear scalar mapper; the third tests descriptor complementarity with the same linear mapper; and the fourth tests the additional nonlinear capacity of the two-layer MLP.

\section{Experimental Results and Analysis}
\label{sec:experiments}
\subsection{Experimental Setup}
\label{subsec:setup}

\subsubsection{Datasets and Evaluation Metrics}
To assess the effectiveness and robustness of the proposed framework, experiments are conducted on two representative multimodal remote sensing datasets:
\begin{itemize}
    \item \textbf{M4-SAR:} A large-scale benchmark for Optical--SAR fusion object detection. It comprises 112,174 instance-level aligned image pairs and approximately one million labeled instances with arbitrary orientations across six categories~\cite{wang2025m4}.
    
    \item \textbf{SpaceNet6-OTD:} A specialized dataset for oriented oil tank detection, constructed from paired high-resolution SAR and optical imagery~\cite{shermeyer2020spacenet, Zhang22otd}. It features densely arranged objects and complex backgrounds, offering a challenging benchmark for assessing cross-modal robustness and generalization.
\end{itemize}
The primary evaluation metric is mean Average Precision (mAP) for oriented bounding box (OBB) detection. We report $\text{mAP}_{50}$ and $\text{mAP}_{50:95}$ under three test settings, namely Full Modality, Masked Optical (SAR only), and Masked SAR (Optical only), to evaluate robustness under different modality availability conditions.

\subsubsection{Implementation Details}
We evaluate one router per detector. Simple U-Net and nnU-Net~\cite{ronneberger2015u,isensee2021nnu} adopt FR since their first cross-modal interaction occurs in the shallow encoder. CSSA, CLANet, and CFT~\cite{23CSSA,he2023multispectral,qingyun2022cross} instead use SR because their dual backbones preserve modality-separated semantic streams until late fusion. FR and SR are not combined in one model.

All models are implemented in PyTorch and trained on NVIDIA RTX 4090 GPUs. Tables~\ref{tab:m4sar_results} and~\ref{tab:spacenet6_results} separate architecture and training effects with four conditions: Baseline/Clean, Router/Clean, Baseline/MD, and Router/MD. The Clean pair isolates router architecture under nominal full-input training; the Router/Clean versus Router/MD difference measures interaction with missing-modality exposure. MD samples the three states in formula~\eqref{eq:training_state_distribution}. Unless otherwise specified, $p=0.2$, giving probabilities $0.6/0.2/0.2$ for full input, Missing Optical, and Missing SAR; the two modalities are not masked together. Optimization is performed with AdamW at an initial learning rate of $10^{-3}$ and with cosine annealing throughout training; all inputs are resized to $512 \times 512$ pixels. Within each controlled comparison, the schedule, augmentation, and modality-mask sampler are identical.

\subsubsection{Budget-Matched CSSA Adaptations of Existing Methods}
For the resource-matched comparison, CSSA serves as the common detector. Every external baseline is a detection adaptation of one existing method, not a direct reproduction of its original task-specific architecture or schedule. SE and CBAM are inserted immediately before the original CSSA fusion module and use no auxiliary loss. For all reported performance entries, we conduct three independent runs and present the mean $\pm$ standard deviation. The remaining adaptation protocols are defined as follows:

\textit{CML-style CSSA}~\cite{ma2023cml} is our detection-specific adaptation of the original classification regularizer. The detector's ground-truth assigner first constructs a positive index set $\mathcal P$ from the full-modality forward pass; the same indices and target classes are then reused for the missing-modality pass so that confidence is compared at identical locations. We identify the native OBB detection score as $s_i=p_i^{obj}p_i(y_i)$ when the head has a separate objectness branch, and as $s_i=p_i(y_i)$ otherwise. The positive-set confidence is $C(x)=|\mathcal P|^{-1}\sum_{i\in\mathcal P}s_i(x)$. For a full input $x^S$ and a sampled subset $x^T$ with $T\subset S$, the objective is
\begin{equation}
\begin{aligned}
\mathcal L_{\mathrm{CML}}&=\max(0,C(x^T)-C(x^S)),\\
\mathcal L&=\mathcal L_{\mathrm{det}}^S+\mathcal L_{\mathrm{det}}^T
+\lambda_{\mathrm{CML}}\mathcal L_{\mathrm{CML}}.
\end{aligned}
\label{eq:cml_cssa}
\end{equation}
where $\mathcal L_{\mathrm{det}}^S$ and $\mathcal L_{\mathrm{det}}^T$ are the detection losses for the full and sampled-subset inputs, respectively, and $\lambda_{\mathrm{CML}}$ weights the CML regularizer. We select $\lambda_{\mathrm{CML}}$ from the pre-declared grid $\{1,5,10,20\}$ on the validation split and train for 300 epochs. Note that this fixed-location proposal formulation is specific to our OBB adaptation and is not the original CML definition.

\textit{ShaSpec-style CSSA}~\cite{wang2023multi} decomposes the last modality-indexed CSSA feature $F_m\in\mathbb R^{C\times H\times W}$ into a shared representation $r_m=E_{\mathrm{sha}}(F_m)$ and a specific representation $s_m=E_m^{\mathrm{spec}}(F_m)$. The weights of $E_{\mathrm{sha}}$ are fully tied between Optical and SAR; the two specific encoders are independent. Each branch outputs $C/2$ channels at this single pre-mixing scale, and $P([r_m,s_m])\in\mathbb R^{C\times H\times W}$. Following the residual form reported in the source method, $\hat F_m=P([r_m,s_m])+F_m$ is passed to the original CSSA fusion module. Its objective is
\begin{equation}
\mathcal L=\mathcal L_{\mathrm{det}}+0.1\mathcal L_{\mathrm{DAO}}+0.02\mathcal L_{\mathrm{DCO}},
\label{eq:shaspec_cssa}
\end{equation}
where DAO uses the source method's $L_1$ distribution-alignment option, and DCO is a modality-domain classification loss that trains a classifier to identify the source modality. Both auxiliary losses are applied only at the last pre-mixing scale, and all added encoders, projections, and classifiers are included in the parameter and FLOP counts. Training follows the common 300-epoch budget.

\textit{SimMLM-style CSSA}~\cite{li2025simmlm} adopts DMoME weighting with MoFe ranking for the two CSSA streams. A gating network produces $w_m=\exp(g_m)/\sum_j\exp(g_j)$, with $g_m=-\infty$ when modality $m$ is absent; the weighted features $\tilde F_m=w_mF_m$ retain the original CSSA fusion module. Given a more-modal input $x^+$ and fewer-modal input $x^-\subset x^+$, the corresponding MoFe objective is
\begin{equation}
\begin{aligned}
\mathcal L_{\mathrm{MoFe}}&=\max(0,\mathcal L_{\mathrm{det}}(x^+)-\mathcal L_{\mathrm{det}}(x^-)),\\
\mathcal L&=\mathcal L_{\mathrm{det}}(x^+)+\mathcal L_{\mathrm{det}}(x^-)
+0.1\mathcal L_{\mathrm{MoFe}}.
\end{aligned}
\label{eq:simmlm_cssa}
\end{equation}
To match the 300-epoch budget, two experts are initialized independently for 50 epochs and cooperatively trained with the gate for 250 epochs. This two-stage allocation is selected under our common optimization budget rather than borrowed from a universal SimMLM schedule.

\textit{M3AE-style CSSA (budget-matched)}~\cite{liu2023m3ae} adds a lightweight reconstruction decoder to the CSSA bottleneck during pretraining and removes it at inference. The 2-D patch-mask ratio is chosen from $r\in\{0.5,0.75,0.875\}$ on the M4-SAR validation split instead of transferring the BraTS-specific value without validation. The first-stage objective is $\mathcal L_{\mathrm{rec}}=\|\hat X-X\|_2^2$. Fine-tuning samples two availability patterns of the same pair and performs latent self-distillation,
\begin{equation}
\mathcal L=\mathcal L_{\mathrm{det}}(x^0)+\mathcal L_{\mathrm{det}}(x^1)+0.1\|f(x^0)-f(x^1)\|_2^2.
\label{eq:m3ae_cssa}
\end{equation}
We allocate 50 epochs to reconstruction pretraining and 250 epochs to detection fine-tuning, preserving the shared optimization budget. This is a resource-matched adaptation rather than a reproduction of the original M3AE schedule, which uses 600 pretraining and 300 fine-tuning epochs on 3-D BraTS data.

\textit{Flex-MoE-style CSSA (budget-matched)}~\cite{yun2024flex} retains the original missing-modality bank, generalized router, specialized router, sparse MoE layer, and combination-specific expert assignment. For the two-modality setting, the valid states are full, Optical-only, and SAR-only, but the expert count and active experts are selected from the reported search spaces $|E|\in\{16,32\}$ and $k\in\{2,3,4\}$ on the validation split rather than collapsed to three experts. A conditional bank supplies $F_m=B_{S,m}$ when $m\notin S$. In this case,
\begin{equation}
\mathcal L=\mathcal L_{\mathrm{det}}+0.01(\mathcal L_{\mathrm{ce}}+\mathcal L_{\mathrm{balance}}),
\label{eq:flexmoe_cssa}
\end{equation}
with the generalized and specialized routing phases implemented within the common 300-epoch optimization budget. Because the source method was developed for medical classification, this remains a CSSA detection adaptation; its full parameter, FLOP, active-expert, GPU-hour, and auxiliary-loss costs are reported rather than hidden by simplifying the expert bank.
\subsection{Quantitative Results}
\label{subsec:quantitative_results}

Quantitative comparisons on M4-SAR and SpaceNet6-OTD are reported in Tables~\ref{tab:m4sar_results} and~\ref{tab:spacenet6_results}. We first evaluate complete modality absence using Full Modality, Masked Optical (SAR only), and Masked SAR (Optical only). Controlled nonzero modality corruptions are evaluated separately in Section~\ref{subsec:controlled_degradation}. Table~\ref{tab:m4sar_results} first lists the controlled results on M4-SAR.

\begin{table*}[!t]
\centering
\caption{Results on M4-SAR under a $2\times2$ architecture/training control. ``Clean'' represents complete pairs only; ``MD'' follows Eq.~\eqref{eq:training_state_distribution}. Representative Clean+Router rows isolate architectural full-input gains, whereas MD+Router versus Clean+Router isolates the interaction with missing-modality training. Green parenthesized values denote absolute mAP gains over the corresponding baseline under the same training protocol.}
\label{tab:m4sar_results}
\scriptsize
\setlength{\tabcolsep}{7.0pt}
\begin{tabular}{@{}l c cccccc c @{}}
\toprule
\multirow{2}{*}{Method} & \multirow{2}{*}{Train} & \multicolumn{2}{c}{Full Modality} & \multicolumn{2}{c}{Masked Optical} & \multicolumn{2}{c}{Masked SAR} & \multirow{2}{*}{Params (M) $\downarrow$} \\
\cmidrule(lr){3-4} \cmidrule(lr){5-6} \cmidrule(lr){7-8}
& & $\text{mAP}_{50}\uparrow$ & $\text{mAP}_{50:95}\uparrow$ & $\text{mAP}_{50}\uparrow$ & $\text{mAP}_{50:95}\uparrow$ & $\text{mAP}_{50}\uparrow$ & $\text{mAP}_{50:95}\uparrow$ & \\
\midrule
SuperYOLO~\cite{zhang2023superyolo} & Clean & 72.4 & 46.9 & 8.9  & 4.1  & 31.8 & 19.6 & 4.85 \\
MCHE~\cite{24MCHE}                   & Clean & 81.2 & 52.8 & 12.8 & 7.5  & 34.9 & 22.8 & 145.81 \\
ICAFusion~\cite{SHEN2023109913}      & Clean & 87.3 & 58.8 & 15.0 & 7.4  & 40.3 & 24.4 & 28.97 \\
MMIDet~\cite{24MMIDet}               & Clean & 87.4 & 57.8 & 15.5 & 7.6  & 34.6 & 24.0 & 53.70 \\
\midrule
Simple U-Net~\cite{ronneberger2015u} & Clean & 58.2 & 31.1 & 0.2 & 0.0 & 22.9 & 12.8 & 2.42 \\
Simple U-Net + FR                     & Clean & 64.1\gain{5.9} & 35.2\gain{4.1} & 5.8\gain{5.6} & 2.1\gain{2.1} & 30.7\gain{7.8} & 16.9\gain{4.1} & 2.44\paraminc{0.02} \\
Simple U-Net                          & MD    & 43.7 & 23.4 & 8.4 & 3.1 & 27.6 & 14.9 & 2.42 \\
Simple U-Net + FR                     & MD    & 63.2\gain{19.5} & 34.7\gain{11.3} & 26.0\gain{17.6} & 10.4\gain{7.3} & 45.6\gain{18.0} & 24.4\gain{9.5} & 2.44\paraminc{0.02} \\
\addlinespace
nnU-Net~\cite{isensee2021nnu}        & Clean & 63.2 & 34.7 & 3.9 & 0.0 & 27.4 & 14.5 & 12.34 \\
nnU-Net                               & MD    & 48.8 & 26.1 & 11.6 & 4.8 & 32.7 & 18.3 & 12.34 \\
nnU-Net + FR                          & MD    & 77.6\gain{28.8} & 50.3\gain{24.2} & 33.7\gain{22.1} & 17.1\gain{12.3} & 52.6\gain{19.9} & 31.5\gain{13.2} & 12.36\paraminc{0.02} \\
\addlinespace
CSSA~\cite{23CSSA}                   & Clean & 83.6 & 54.1 & 13.1 & 8.6 & 28.8 & 20.2 & 13.51 \\
CSSA + SR                             & Clean & 85.4\gain{1.8} & 55.8\gain{1.7} & 15.6\gain{2.5} & 9.4\gain{0.8} & 34.7\gain{5.9} & 22.9\gain{2.7} & 14.03\paraminc{0.52} \\
CSSA                                  & MD    & 62.7 & 40.6 & 18.7 & 10.4 & 38.2 & 22.6 & 13.51 \\
CSSA + SR                             & MD    & 83.7\gain{21.0} & 54.7\gain{14.1} & 26.3\gain{7.6} & 14.6\gain{4.2} & 55.7\gain{17.5} & 32.1\gain{9.5} & 14.03\paraminc{0.52} \\
\addlinespace
CLANet~\cite{he2023multispectral}    & Clean & 68.3 & 51.6 & 11.7 & 6.3 & 30.6 & 21.9 & 48.21 \\
CLANet                               & MD    & 55.4 & 40.8 & 17.1 & 9.5 & 37.9 & 23.8 & 48.21 \\
CLANet + SR                          & MD    & 79.8\gain{24.4} & 53.6\gain{12.8} & 25.9\gain{8.8} & 15.8\gain{6.3} & 52.7\gain{14.8} & 31.2\gain{7.4} & 48.73\paraminc{0.52} \\
\addlinespace
CFT~\cite{qingyun2022cross}          & Clean & 87.6 & 59.0 & 17.9 & 10.4 & 35.3 & 22.7 & 53.76 \\
CFT + SR                              & Clean & 89.1\gain{1.5} & 60.3\gain{1.3} & 20.6\gain{2.7} & 12.1\gain{1.7} & 39.4\gain{4.1} & 25.1\gain{2.4} & 54.28\paraminc{0.52} \\
CFT                                   & MD    & 74.6 & 48.3 & 22.4 & 12.7 & 42.8 & 26.5 & 53.76 \\
CFT + SR                              & MD    & 87.8\gain{13.2} & 59.2\gain{10.9} & 31.7\gain{9.3} & 20.0\gain{7.3} & 57.3\gain{14.5} & 36.2\gain{9.7} & 54.28\paraminc{0.52} \\
\bottomrule
\end{tabular}
\end{table*}

\subsubsection{Performance on M4-SAR}
Representative multimodal detectors such as ICAFusion and MMIDet achieve competitive performance under full-modality inputs, but their accuracy drops substantially when one modality is absent. For example, MMIDet's $\text{mAP}_{50}$ decreases from 87.4\% to 15.5\% under the SAR-only setting.

With the controlled MD protocol, FR improves Simple U-Net from 43.7\% to 63.2\% under full input, from 8.4\% to 26.0\% under Masked Optical, and from 27.6\% to 45.6\% under Masked SAR. SR raises CSSA from 62.7\% to 83.7\%, from 18.7\% to 26.3\%, and from 38.2\% to 55.7\%, respectively. The nnU-Net, CLANet, and CFT blocks follow the same comparison structure and exhibit consistent gains under both missing directions.

Table~\ref{tab:spacenet6_results} next gives the corresponding comparison on SpaceNet6-OTD.

\begin{table*}[ht]
\centering
\caption{Results on SpaceNet6-OTD with the same architecture/training controls as Table~\ref{tab:m4sar_results}. Clean+Router controls are reported for Simple U-Net, CSSA, and CFT. Green parenthesized values denote absolute mAP gains over the corresponding baseline under the same training protocol.}
\label{tab:spacenet6_results}
\scriptsize
\setlength{\tabcolsep}{7.0pt}
\begin{tabular}{@{}l c cccccc c @{}}
\toprule
\multirow{2}{*}{Method} & \multirow{2}{*}{Train} & \multicolumn{2}{c}{Full Modality} & \multicolumn{2}{c}{Masked Optical} & \multicolumn{2}{c}{Masked SAR} & \multirow{2}{*}{Params (M) $\downarrow$} \\
\cmidrule(lr){3-4} \cmidrule(lr){5-6} \cmidrule(lr){7-8}
& & $\text{mAP}_{50}\uparrow$ & $\text{mAP}_{50:95}\uparrow$ & $\text{mAP}_{50}\uparrow$ & $\text{mAP}_{50:95}\uparrow$ & $\text{mAP}_{50}\uparrow$ & $\text{mAP}_{50:95}\uparrow$ & \\
\midrule
SuperYOLO~\cite{zhang2023superyolo} & Clean & 96.9 & 84.1 & 9.7 & 5.4 & 77.8 & 55.1 & 4.85 \\
MCHE~\cite{24MCHE}                   & Clean & 97.2 & 85.8 & 13.6 & 7.9 & 80.6 & 57.9 & 145.81 \\
ICAFusion~\cite{SHEN2023109913}      & Clean & 96.6 & 82.8 & 6.8 & 3.7 & 74.6 & 52.3 & 28.97 \\
MMIDet~\cite{24MMIDet}               & Clean & 97.4 & 86.5 & 16.9 & 9.8 & 82.4 & 59.3 & 53.70 \\
\midrule
Simple U-Net~\cite{ronneberger2015u} & Clean & 96.2 & 74.2 & 0.0 & 0.0 & 65.8 & 46.2 & 2.42 \\
Simple U-Net + FR                          & Clean & 97.0\gain{0.8} & 76.2\gain{2.0} & 8.7\gain{8.7} & 4.1\gain{4.1} & 70.4\gain{4.6} & 50.3\gain{4.1} & 2.44\paraminc{0.02} \\
Simple U-Net                               & MD    & 92.8 & 70.1 & 25.4 & 12.9 & 82.6 & 58.8 & 2.42 \\
Simple U-Net + FR                          & MD    & 96.3\gain{3.5} & 74.9\gain{4.8} & 61.6\gain{36.2} & 32.2\gain{19.3} & 96.0\gain{13.4} & 74.8\gain{16.0} & 2.44\paraminc{0.02} \\
\addlinespace
nnU-Net~\cite{isensee2021nnu}       & Clean & 96.3 & 77.3 & 0.4 & 0.0 & 66.2 & 50.4 & 12.34 \\
nnU-Net                                     & MD    & 93.6 & 73.5 & 30.8 & 15.7 & 83.7 & 61.2 & 12.34 \\
nnU-Net + FR                                & MD    & 97.1\gain{3.5} & 81.2\gain{7.7} & 72.4\gain{41.6} & 40.6\gain{24.9} & 92.2\gain{8.5} & 69.8\gain{8.6} & 12.36\paraminc{0.02} \\
\addlinespace
CSSA~\cite{23CSSA}                    & Clean & 97.0 & 85.7 & 14.0 & 8.2 & 78.0 & 55.6 & 13.51 \\
CSSA + SR                                   & Clean & 97.6\gain{0.6} & 86.8\gain{1.1} & 22.7\gain{8.7} & 12.6\gain{4.4} & 82.1\gain{4.1} & 59.2\gain{3.6} & 14.03\paraminc{0.52} \\
CSSA                                        & MD    & 94.1 & 81.8 & 35.6 & 19.2 & 86.4 & 63.8 & 13.51 \\
CSSA + SR                                   & MD    & 97.0\gain{2.9} & 86.2\gain{4.4} & 72.8\gain{37.2} & 40.4\gain{21.2} & 96.2\gain{9.8} & 76.0\gain{12.2} & 14.03\paraminc{0.52} \\
\addlinespace
CLANet~\cite{he2023multispectral}     & Clean & 96.9 & 85.4 & 12.3 & 7.2 & 79.2 & 56.7 & 48.21 \\
CLANet                                      & MD    & 93.8 & 81.2 & 33.8 & 18.1 & 86.9 & 64.1 & 48.21 \\
CLANet + SR                                 & MD    & 96.9\gain{3.1} & 86.0\gain{4.8} & 71.0\gain{37.2} & 39.2\gain{21.1} & 96.9\gain{10.0} & 75.8\gain{11.7} & 48.73\paraminc{0.52} \\
\addlinespace
CFT~\cite{qingyun2022cross}           & Clean & 97.5 & 86.9 & 19.8 & 11.6 & 83.6 & 60.4 & 53.76 \\
CFT + SR                                    & Clean & 98.0\gain{0.5} & 87.9\gain{1.0} & 27.6\gain{7.8} & 15.8\gain{4.2} & 87.0\gain{3.4} & 63.1\gain{2.7} & 54.28\paraminc{0.52} \\
CFT                                         & MD    & 94.8 & 83.3 & 41.5 & 22.8 & 88.7 & 66.0 & 53.76 \\
CFT + SR                                    & MD    & 97.6\gain{2.8} & 87.5\gain{4.2} & 79.2\gain{37.7} & 45.5\gain{22.7} & 97.1\gain{8.4} & 77.1\gain{11.1} & 54.28\paraminc{0.52} \\
\bottomrule
\end{tabular}
\end{table*}

\subsubsection{Performance on SpaceNet6-OTD}
On SpaceNet6-OTD, the matched-training results show the same trend: relative to MD-only training, FR enhances Masked-Optical $\text{mAP}_{50}$ from 25.4\% to 61.6\% for Simple U-Net and from 30.8\% to 72.4\% for nnU-Net. SR raises CSSA from 35.6\% to 72.8\%, CLANet from 33.8\% to 71.0\%, and CFT from 41.5\% to 79.2\%. Full-input changes are somewhat smaller on this dataset, consistent with performance saturation near 97\%.

\subsubsection{Parameter Efficiency}
Despite the substantial robustness improvements, the proposed modules incur only marginal parameter overhead. As shown in the ``Params'' column, FR adds only 0.02M parameters to the U-Net-based baselines. Similarly, SR increases the parameter count of dual-backbone models (e.g., CSSA and CFT) by about 0.52M, which remains small relative to the overall model size. These results demonstrate that the proposed framework achieves a favorable trade-off between robustness and model complexity.

\subsubsection{Full-Input Fusion Enhancement}
The Clean+Router controls test whether routing helps without missing-modality exposure. On M4-SAR, FR increases the full-input $\text{mAP}_{50}$ of Simple U-Net from 58.2\% to 64.1\%; SR raises CSSA from 83.6\% to 85.4\% and CFT from 87.6\% to 89.1\%. The corresponding SpaceNet6-OTD gains are smaller because the baselines are already near saturation, but remain positive. These improvements cannot be attributed to modality dropout. They support the hypothesis that nominal full pairs contain sample-dependent conflicts that can be attenuated before feature-value mixing. MD then provides a different benefit: comparing Router/MD with Router/Clean shows large improvements under missing inputs, with a modest full-input trade-off.

\subsection{Ablation Study}
\label{subsec:ablation}

To isolate architectural contributions from missing-modality exposure, every variant in Table~\ref{tab:ablation} is trained with the same modality-dropout probability ($p=0.2$), optimizer, schedule, modality-mask sampler, and augmentation pipeline. Thus, each block changes only the indicated routing component.

\begin{table*}[t]
\centering
\caption{Component ablation on M4-SAR under fixed modality dropout ($p=0.2$). All variants within each block use identical optimization, modality-mask sampling, and data augmentation.}
\label{tab:ablation}
\begin{tabular}{@{}l cccccc @{}}
\toprule
\multirow{2}{*}{Configuration} & \multicolumn{2}{c}{Full Modality} & \multicolumn{2}{c}{\makecell{Masked Optical \\ (SAR only)}} & \multicolumn{2}{c}{\makecell{Masked SAR \\ (Optical only)}} \\
\cmidrule(lr){2-3} \cmidrule(lr){4-5} \cmidrule(lr){6-7}
& $\text{mAP}_{50}\uparrow$ & $\text{mAP}_{50:95}\uparrow$ & $\text{mAP}_{50}\uparrow$ & $\text{mAP}_{50:95}\uparrow$ & $\text{mAP}_{50}\uparrow$ & $\text{mAP}_{50:95}\uparrow$ \\
\midrule
\multicolumn{7}{@{}l}{\textbf{(a) Ablation on FR with U-Net Baseline}} \\
\midrule
Baseline + modality dropout        & 43.7 & 23.4 & 8.4  & 3.1  & 27.6 & 14.9 \\
\quad + Independent scalar routing & 46.1 & 25.0 & 13.7 & 5.4  & 31.8 & 17.2 \\
\quad + Channel selection          & 54.8 & 29.6 & 19.7 & 7.7  & 39.6 & 21.5 \\
\quad + Spatial gating (Full FR)   & 63.2 & 34.7 & 26.0 & 10.4 & 45.6 & 24.4 \\
\midrule
\multicolumn{7}{@{}l}{\textbf{(b) Ablation on SR with CSSA Baseline}} \\
\midrule
Baseline + modality dropout        & 62.7 & 40.6 & 18.7 & 10.4 & 38.2 & 22.6 \\
\quad + Avg descriptor + linear gate       & 68.4 & 44.2 & 21.1 & 11.7 & 43.5 & 25.4 \\
\quad + Max descriptor + linear gate       & 69.2 & 44.8 & 22.5 & 12.4 & 46.1 & 27.0 \\
\quad + Avg--Max descriptor + linear gate  & 75.8 & 49.3 & 24.2 & 13.5 & 50.7 & 29.4 \\
\quad + Avg--Max descriptor + two-layer MLP gate (Full SR) & 83.7 & 54.7 & 26.3 & 14.6 & 55.7 & 32.1 \\
\bottomrule
\end{tabular}
\end{table*}

\subsubsection{Effectiveness of the FR Stem}
Table~\ref{tab:ablation}(a) presents the stepwise FR experiment under a shared training protocol. Under Masked Optical, the MD-only baseline obtains 8.4\% $\text{mAP}_{50}$; scalar routing, channel selection, and spatial gating raise it to 13.7\%, 19.7\%, and 26.0\%, respectively. The monotonic endpoint-consistent trend indicates complementary contributions from the three stages.

\subsubsection{Effectiveness of SR}
Table~\ref{tab:ablation}(b) applies the same control to CSSA using the executable gates in Eqs.~\eqref{eq:sr_ablation_gates}--\eqref{eq:sr_ablation_output}. Under Masked SAR, average-only and maximum-only linear gates attain 43.5\% and 46.1\% $\text{mAP}_{50}$, the dual-statistic linear gate reaches 50.7\%, and the two-layer MLP achieves 55.7\%. This ordering separates descriptor complementarity from nonlinear mapping capacity.

\subsubsection{Controlled Utility-Exposure Comparison}
Table~\ref{tab:fair_routing_comparison} fixes the schedule, optimizer, augmentation, masking sampler, and modality-dropout probability across all MD-trained variants. SE and CBAM remain close to the MD-only baseline, while a modality-level scalar gate yields a larger gain but still underperforms Full FR, which achieves the strongest result in all three input conditions. The shared curriculum prevents unequal low-utility exposure from explaining these differences.

\begin{table*}[t]
\caption{Controlled comparison on M4-SAR. All MD-trained variants use the same 300 epochs, optimizer, modality-mask sampler, augmentation, and modality-dropout probability ($p=0.2$). The clean baseline is diagnostic only.}
\label{tab:fair_routing_comparison}
\centering
\setlength{\tabcolsep}{3.7pt}
\begin{tabular}{@{}l cccccc c@{}}
\toprule
\multirow{2}{*}{Method} & \multicolumn{2}{c}{Full Modality} & \multicolumn{2}{c}{Masked Optical} & \multicolumn{2}{c}{Masked SAR} & Params $\downarrow$ \\
\cmidrule(lr){2-3}\cmidrule(lr){4-5}\cmidrule(lr){6-7}
& mAP50 $\uparrow$ & mAP50:95 $\uparrow$ & mAP50 $\uparrow$ & mAP50:95 $\uparrow$ & mAP50 $\uparrow$ & mAP50:95 $\uparrow$ & (M) \\
\midrule
Baseline (clean training)       & 58.2 & 31.1 & 0.2  & 0.0 & 22.9 & 12.8 & 2.42 \\
Baseline + modality dropout     & 43.7 & 23.4 & 8.4  & 3.1 & 27.6 & 14.9 & 2.42 \\
Baseline + SE                   & 44.0 & 23.7 & 9.1  & 3.5 & 28.1 & 15.2 & 2.43 \\
Baseline + CBAM                 & 44.2 & 23.8 & 9.5  & 3.7 & 28.4 & 15.4 & 2.43 \\
Baseline + scalar gate          & 46.1 & 25.0 & 13.7 & 5.4 & 31.8 & 17.2 & 2.42 \\
Baseline + FR (ours)            & \textbf{63.2} & \textbf{34.7} & \textbf{26.0} & \textbf{10.4} & \textbf{45.6} & \textbf{24.4} & 2.44 \\
\bottomrule
\end{tabular}
\end{table*}

\subsubsection{Budget-Matched Adaptations of Existing Methods}
Table~\ref{tab:robust_fusion_comparison} compares CSSA detection adaptations inspired by CML, ShaSpec, M3AE, Flex-MoE, and SimMLM. The original schedules and task-specific architectures are not involved in these direct reproductions: the M3AE-style row uses a reduced 50+250 budget, the SimMLM-style allocation follows the common budget, and the Flex-MoE-style row retains the bank/router/sparse-MoE components while searching the reported expert and top-$k$ ranges. Under this budget, the external adaptations improve on CSSA+MD with different computation costs, while SR reaches the highest three $\text{mAP}_{50}$ values with substantially lower training cost than the reconstruction, two-stage, and sparse-MoE adaptations.

\begin{table*}[t]
\caption{Budget-matched CSSA adaptations inspired by existing methods on M4-SAR. These are re-implemented detection adaptations rather than direct reproductions of the original task-specific architectures. All rows use the same modality-dropout sampler and 300-epoch optimization budget. Performance is reported as mean $\pm$ standard deviation over three independent runs; Params and GFLOPs denote inference-time cost at $512\times512$.}
\label{tab:robust_fusion_comparison}
\centering
\scriptsize
\setlength{\tabcolsep}{2.6pt}
\begin{tabular}{@{}>{\raggedright\arraybackslash}p{2.20cm} >{\raggedright\arraybackslash}p{2.70cm} c c c >{\raggedright\arraybackslash}p{2.65cm} ccc@{}}
\toprule
Method & Adaptation & Params (M) $\downarrow$ & GFLOPs $\downarrow$ & GPU-h $\downarrow$ & Auxiliary objective & \makecell{Full\\mAP50 $\uparrow$} & \makecell{Miss OPT\\mAP50 $\uparrow$} & \makecell{Miss SAR\\mAP50 $\uparrow$} \\
\midrule
\multicolumn{9}{@{}l}{\textit{\textbf{CSSA}}} \\
\addlinespace[1pt]
+ MD & None (MD only) & 13.51 & 42.6 & 18.4 & none & 62.7$\pm$0.4 & 18.7$\pm$0.7 & 38.2$\pm$0.6 \\
+ SE~\cite{hu2018squeeze} & Channel recalibration & 13.54 & 42.7 & 19.0 & none & 63.4$\pm$0.5 & 19.8$\pm$0.6 & 39.6$\pm$0.7 \\
+ CBAM~\cite{woo2018cbam} & Channel--spatial attention & 13.56 & 42.9 & 19.5 & none & 64.0$\pm$0.4 & 20.5$\pm$0.8 & 39.2$\pm$0.5 \\
+ CML~\cite{ma2023cml} & Confidence ranking & 13.51 & 42.6 & 32.8 & $\lambda_{\mathrm{CML}}L_{\mathrm{CML}}$ & 66.1$\pm$0.6 & 22.8$\pm$0.7 & 44.5$\pm$0.8 \\
+ ShaSpec~\cite{wang2023multi} & Shared--specific features & 15.20 & 45.8 & 26.7 & $0.1L_{\mathrm{DAO}}+0.02L_{\mathrm{DCO}}$ & 70.2$\pm$0.5 & 23.9$\pm$0.9 & 47.1$\pm$0.6 \\
+ M3AE~\cite{liu2023m3ae} & Recon. + distillation & 15.42 & 43.9 & 38.9 & $L_{\mathrm{rec}}+0.1L_{\mathrm{con}}$ & 71.5$\pm$0.8 & 24.9$\pm$0.6 & 49.3$\pm$0.9 \\
+ Flex-MoE~\cite{yun2024flex} & Sparse MoE routing & 31.86 & 48.7 & 44.6 & $0.01(L_{\mathrm{ce}}+L_{\mathrm{balance}})$ & 73.6$\pm$0.7 & 25.7$\pm$0.8 & 50.8$\pm$0.7 \\
+ SimMLM~\cite{li2025simmlm} & DMoME + MoFe & 14.06 & 43.1 & 34.2 & $0.1L_{\mathrm{MoFe}}$ & 75.9$\pm$0.5 & 25.4$\pm$0.6 & 53.1$\pm$0.8 \\
+ SR (ours) & Boundary routing & 14.03 & 42.9 & 19.7 & none & 83.7$\pm$0.3 & 26.3$\pm$0.5 & 55.7$\pm$0.6 \\
\bottomrule
\end{tabular}
\end{table*}

\subsubsection{Effect of Routing Position}
Table~\ref{tab:routing_position} reports a controlled study in which router capacity and training are held approximately fixed while the gate is moved through CSSA. Routing immediately before the first learned cross-modal feature-value mixing operation performs best across full and missing inputs. Moving the same-capacity gate after feature-value mixing leads to a marked drop, supporting the boundary-alignment hypothesis rather than a generic benefit from adding parameters.

\begin{table*}[t]
\caption{Routing-position study under a matched parameter budget on M4-SAR. The same router is moved through CSSA without changing the modality-mask curriculum. ``Before feature-value mixing'' denotes placement immediately before the first learned cross-modal feature-value mixing boundary.}
\label{tab:routing_position}
\centering
\setlength{\tabcolsep}{4.5pt}
\begin{tabular}{@{}l cccccc c@{}}
\toprule
\multirow{2}{*}{Routing position} & \multicolumn{2}{c}{Full Modality} & \multicolumn{2}{c}{Masked Optical} & \multicolumn{2}{c}{Masked SAR} & Params $\downarrow$ \\
\cmidrule(lr){2-3}\cmidrule(lr){4-5}\cmidrule(lr){6-7}
& mAP50 $\uparrow$ & mAP50:95 $\uparrow$ & mAP50 $\uparrow$ & mAP50:95 $\uparrow$ & mAP50 $\uparrow$ & mAP50:95 $\uparrow$ & (M) \\
\midrule
Raw input                          & 78.1 & 50.2 & 21.4 & 11.8 & 45.0 & 26.3 & 14.02 \\
After first layer                  & 80.0 & 51.7 & 23.6 & 12.9 & 48.2 & 28.1 & 14.03 \\
Before feature-value mixing (ours) & \textbf{83.7} & \textbf{54.7} & \textbf{26.3} & \textbf{14.6} & \textbf{55.7} & \textbf{32.1} & 14.03 \\
After feature-value mixing         & 77.2 & 49.4 & 18.8 & 10.1 & 42.3 & 24.7 & 14.03 \\
Before late semantic mixing        & 75.6 & 48.1 & 17.2 & 9.3  & 39.7 & 23.0 & 14.04 \\
Before detection head             & 73.9 & 46.8 & 15.9 & 8.5  & 36.8 & 21.4 & 14.03 \\
\bottomrule
\end{tabular}
\end{table*}

\subsubsection{Sensitivity to Modality Dropout Probability}
\setcounter{topnumber}{1}
Table~\ref{tab:dropout_p_sensitivity} evaluates $p\in\{0,0.1,0.2,0.3,0.5\}$. Robustness is improved for $p=0.2$, while $p=0.3$ yields only small missing-modality gains and reduces full-input accuracy; $p=0.5$ collapses under the stated schedule. We therefore choose $p=0.2$ as the clean--robustness trade-off.

\begin{table}[!t]
\caption{Sensitivity of the Modality Dropout probability $p$ on M4-SAR using Simple U-Net with FR. ``Collapse'' indicates failure to converge under the stated training schedule.}
\label{tab:dropout_p_sensitivity}
\centering
\scriptsize
\setlength{\tabcolsep}{1.5pt}
\renewcommand{\arraystretch}{1.1}
\begin{tabular}{@{}c cccccc@{}}
\toprule
\multirow{2}{*}{$p$} & \multicolumn{2}{c}{Full Modality} & \multicolumn{2}{c}{Masked Optical} & \multicolumn{2}{c}{Masked SAR} \\
\cmidrule(lr){2-3} \cmidrule(lr){4-5} \cmidrule(lr){6-7}
& mAP50 $\uparrow$ & mAP50:95 $\uparrow$ & mAP50 $\uparrow$ & mAP50:95 $\uparrow$ & mAP50 $\uparrow$ & mAP50:95 $\uparrow$ \\
\midrule
0.0 & 64.1 & 35.2 & 5.8  & 2.1  & 30.7 & 16.9 \\
0.1 & 64.0 & 35.0 & 18.4 & 7.4  & 39.2 & 21.1 \\
0.2 & 63.2 & 34.7 & 26.0 & 10.4 & 45.6 & 24.4 \\
0.3 & 56.8 & 29.6 & 26.4 & 10.6 & 46.1 & 24.8 \\
0.5 & \multicolumn{6}{c}{Collapse} \\
\bottomrule
\end{tabular}
\end{table}

\subsection{Routing-Weight Response to Modality Availability}
\label{subsec:routing_behavior}
Table~\ref{tab:routing_behavior} summarizes two separate routing experiments. FR and SR retain high weights for available streams and reduce the masked stream. Since this test has zero-valued inputs, it demonstrates availability awareness only; it neither calibrates the weights as probabilities nor establishes reliability estimation under nonzero corruption.

\begin{table}[!t]
\caption{Task-conditioned routing weights from two separate architecture-specific experiments: FR ($w$) in Simple U-Net and SR ($\alpha$) in CSSA. Values are mean $\pm$ standard deviation over each test set; FR and SR are not used jointly.}
\label{tab:routing_behavior}
\centering
\scriptsize
\setlength{\tabcolsep}{1.2mm}
\begin{tabular}{@{}l cccc @{}}
\toprule
\multirow{2}{*}{Scenario} & \multicolumn{2}{c}{FR Scalar ($w$)} & \multicolumn{2}{c}{SR Gate ($\alpha$)} \\
\cmidrule(lr){2-3} \cmidrule(lr){4-5}
& OPT ($w_{opt}$) & SAR ($w_{sar}$) & OPT ($\alpha_{opt}$) & SAR ($\alpha_{sar}$) \\
\midrule
Full Modality & 0.81$\pm$0.07 & 0.78$\pm$0.09 & 0.92$\pm$0.03 & 0.89$\pm$0.05 \\
Masked Optical & 0.15$\pm$0.06 & 0.84$\pm$0.08 & 0.11$\pm$0.07 & 0.91$\pm$0.07 \\
Masked SAR & 0.86$\pm$0.07 & 0.17$\pm$0.07 & 0.93$\pm$0.02 & 0.16$\pm$0.03 \\
\bottomrule
\end{tabular}
\end{table}

\subsection{Task Utility and Negative Transfer}
\label{subsec:utility_analysis}
Table~\ref{tab:shift_utility} isolates cross-modal correspondence from intrinsic image appearance by circularly translating SAR by $\{0,2,4,8,16\}$ pixels. Circular translation preserves each channel's pixel distribution exactly, while progressively disrupting Optical--SAR correspondence. For FR, $\bar r_m$ averages the complete multiplicative transmission assigned to modality-indexed group $m$ after scalar, channel, and spatial modulation. As displacement increases, both detectors lose accuracy, but FR degrades more slowly and reduces SAR transmission from 0.78 to 0.39. This behavior links the gate to joint task contribution rather than visible image quality.

We further compare learned weights with the model-specific LOMO utility in Eq.~\eqref{eq:oracle_utility}. Table~\ref{tab:utility_correlation} uses MD-trained frozen checkpoints and reports positive Spearman correlations for both modalities and both router types. This evidence supports the utility interpretation, although it is not guaranteed by the model design. FR exhibits stronger correlations because its cross-conditioned channel/spatial gates directly observe the paired representation; SR remains a weaker stream-level proxy based only on modality-specific statistics. We therefore do not view SR as an explicit misalignment or conflict estimator.

\begin{figure*}[!t]
\centering
\includegraphics[width=0.90\textwidth]{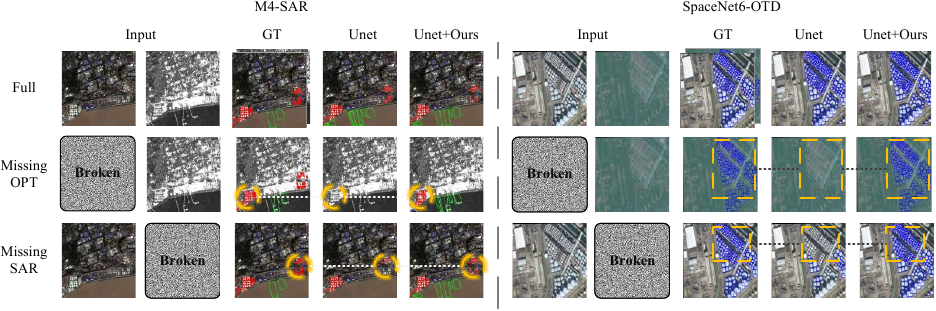}
\caption{Qualitative comparison under full- and missing-modality settings. ``Broken'' denotes complete input unavailability. Yellow circles and arrows highlight baseline errors mitigated by task-conditioned routing.}
\label{fig:qualitative_results}
\end{figure*}

\begin{table*}[!t]
\centering
\begin{minipage}[t]{0.49\textwidth}
\centering
\captionof{table}{M4-SAR correspondence-shift control (Baseline/+FR: $\mathrm{mAP}_{50}$). Circular translation preserves per-channel pixel distributions; $\bar r_m$ denotes mean FR transmission.}
\label{tab:shift_utility}
\footnotesize
\setlength{\tabcolsep}{4pt}
\begin{tabular}{@{}c cc cc@{}}
\toprule
Shift (px) & Baseline $\uparrow$ & +FR $\uparrow$ & $\bar r_{opt}$ & $\bar r_{sar}$ \\
\midrule
0  & 58.2 & 64.1 & 0.81 & 0.78 \\
2  & 57.5 & 63.5 & 0.82 & 0.74 \\
4  & 55.8 & 62.0 & 0.83 & 0.68 \\
8  & 51.6 & 59.3 & 0.84 & 0.57 \\
16 & 43.2 & 53.8 & 0.86 & 0.39 \\
\bottomrule
\end{tabular}
\end{minipage}\hfill
\begin{minipage}[t]{0.49\textwidth}
\centering
\captionof{table}{Spearman correlation between routing weights and model-specific LOMO utility. All listed correlations are significant at $p<0.001$.}
\label{tab:utility_correlation}
\footnotesize
\setlength{\tabcolsep}{4.5pt}
\begin{tabular}{@{}l c cc@{}}
\toprule
Router & Modality & M4-SAR & SpaceNet6-OTD \\
\midrule
FR & Optical & 0.62 & 0.57 \\
FR & SAR     & 0.66 & 0.61 \\
SR & Optical & 0.49 & 0.45 \\
SR & SAR     & 0.53 & 0.48 \\
\bottomrule
\end{tabular}
\end{minipage}
\end{table*}

\begin{table}[!t]
\caption{Negative-transfer rate (NTR, \%) on M4-SAR. Lower is better; ``Base'' and ``+FR/SR'' use matched protocols. ``+FR/SR'' denotes FR for Simple U-Net and SR for CSSA and CFT.}
\label{tab:ntr_results}
\centering
\footnotesize
\setlength{\tabcolsep}{2.4pt}
\begin{tabular}{@{}l c ccc@{}}
\toprule
Backbone & Train & Base $\downarrow$ & +FR/SR $\downarrow$ & $\Delta$ (pp) \\
\midrule
Simple U-Net & Clean & 31.8 & 20.6 & $-11.2$ \\
Simple U-Net & MD    & 29.4 & 16.7 & $-12.7$ \\
CSSA         & Clean & 27.5 & 18.2 & $-9.3$ \\
CSSA         & MD    & 25.8 & 14.9 & $-10.9$ \\
CFT          & Clean & 23.1 & 15.6 & $-7.5$ \\
CFT          & MD    & 21.4 & 12.8 & $-8.6$ \\
\bottomrule
\end{tabular}
\end{table}

Finally, Table~\ref{tab:ntr_results} evaluates the sample-level failure mode defined in Eq.~\eqref{eq:ntr}. The MD rows serve as the primary comparison because their checkpoints are trained on the same three input configurations evaluated by NTR: full, Optical-masked (SAR-only), and SAR-masked (Optical-only). Clean rows are diagnostic only: their zero-stream counterfactuals are out of distribution and may conflate modality contribution with missing-input fragility. Across both protocols, the matched FR/SR variants consistently reduce NTR. Together, the clean-router, utility-correlation, and NTR controls support negative-transfer mitigation across nominal, corrupted, and missing-input settings without treating LOMO utility as modality-intrinsic.

\subsection{Qualitative Analysis}
\label{subsec:qualitative_analysis}
Fig.~\ref{fig:qualitative_results} exhibits detection results under full- and missing-modality settings. Compared with the baseline, the proposed framework produces fewer false alarms and missed detections, indicating that unavailable or low-utility stream responses are attenuated under incomplete inputs.

\subsection{Robustness under Controlled Modality Corruptions}
\label{subsec:controlled_degradation}
\begin{figure}[!t]
\centering
\includegraphics[width=\columnwidth]{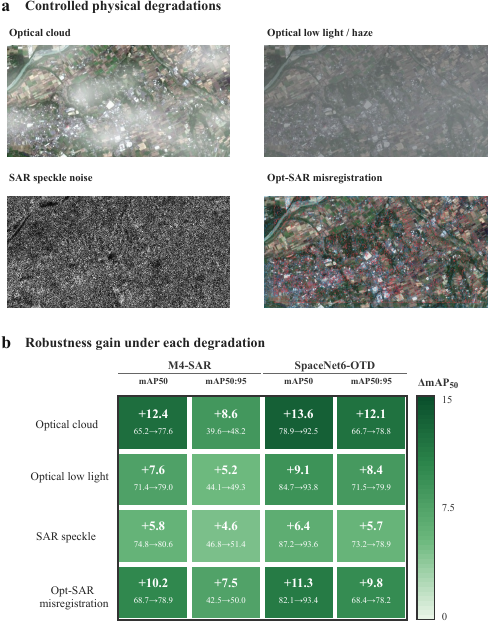}
\caption{Controlled low-utility conditions. 
(a) Visual examples of optical cloud, optical low-light/haze, SAR speckle, and composite SAR translation with boundary corruption. 
(b) Detection performance of CFT+SR against CFT, with absolute gains reported under each condition. The misregistration panel is an additional observation rather than evidence that SR explicitly estimates geometric alignment.}
\label{fig:controlled_degradation}
\end{figure}

Beyond complete modality absence, we further evaluate the four controlled conditions in Fig.~\ref{fig:controlled_degradation}: optical cloud occlusion, optical low-light/haze, SAR speckle noise, and Optical--SAR misregistration. Cloud occlusion is synthesized by alpha-blending blurred anisotropic bright masks with the Optical image. Low-light/haze combines intensity attenuation, contrast compression, and a spatially varying gray veil. SAR speckle uses multiplicative Gamma noise with a small additive Gaussian component. The displayed misregistration example translates an edge-enhanced SAR rendering relative to Optical, with cropping and boundary fill; it is therefore distinct from the distribution-preserving circular shifts in Table~\ref{tab:shift_utility}.

Fig.~\ref{fig:controlled_degradation}(b) reports CFT+SR against CFT under each condition. Cloud, low-light/haze, and speckle test nonzero changes to modality content. Misregistration is retained only as an additional observation: SR can attenuate a stream that becomes less useful under this composite shift-and-boundary corruption, but it does not explicitly estimate geometric alignment since its independent global-statistic gates contain no cross-modal correspondence term.

\section{Conclusion}
\label{sec:conclusion}
This work investigates Optical--SAR fusion from a model-specific task-utility perspective under imperfect correspondence and learns contribution routing from detection supervision alone. We define the first learned cross-modal feature-value mixing boundary and deploy one topology-aligned router per detector: FR enables cross-modal interaction in the control path while preserving source-indexed value paths near an early boundary, whereas SR predicts a stream-level contribution weight from modality-specific statistics before late fusion.

\setcounter{topnumber}{2}

Detection supervision produces task-conditioned weights; their positive correlation with model-specific LOMO utility provides empirical support for the utility interpretation without turning them into calibrated reliability, confidence, or uncertainty estimates. LOMO utility is not employed as an optimization target, yet it provides an analysis-only reference for examining whether the learned weights track model-specific contribution. Clean-input controls show that routing can improve ordinary full-modality fusion by reducing negative transfer, while controlled shifts and NTR analysis further connect the gains to task contribution. Missing modalities and nonzero corruptions are treated as extreme low-utility conditions, and SR remains a stream-level proxy rather than an explicit geometric-alignment estimator. The approach incurs limited overhead and requires no reconstruction or distillation objective.

\balance
\bibliographystyle{IEEEtran}
\bibliography{main}
\end{document}